\documentclass{article}

\usepackage{PRIMEarxiv}

\usepackage[utf8]{inputenc} 
\usepackage[T1]{fontenc}    
\usepackage{hyperref}       
\usepackage{url}            
\usepackage{booktabs}       
\usepackage{amsfonts}       
\usepackage{nicefrac}       
\usepackage{microtype}      
\usepackage{lipsum}
\usepackage{changepage}
\usepackage{fancyhdr}       
\usepackage{graphicx}       
\graphicspath{{media/}}     
\usepackage[
    backend=biber,
    style=numeric,
    sorting=none
  ]{biblatex}
\title{PoxVerifi: An Information Verification System to Combat Monkeypox Misinformation
}

\author{
  Akaash Kolluri\\
  Computational Media Lab\\
  The University of Texas at Austin\\
  \texttt{akaashrkolluri@computationalmedialab.com} \\
   \And
  Kami Vinton \\
  School of Journalism and Media \\
  The University of Texas at Austin \\
  \texttt{kamivinton@utexas.edu} \\
  \And
   Dhiraj Murthy \\
   Computational Media Lab\\
School of Journalism and Media\\
   The University of Texas at Austin \\
   \texttt{Dhiraj.Murthy@austin.utexas.edu} \\
}

\begin{document}
\maketitle

\begin{abstract}
Following recent outbreaks, monkeypox-related misinformation continues to rapidly spread online. This negatively impacts response strategies and disproportionately harms LGBTQ+ communities in the short-term, and ultimately undermines the overall effectiveness of public health responses. In an attempt to combat monkeypox-related misinformation, we present PoxVerifi, an open-source, extensible tool that provides a comprehensive approach to assessing the accuracy of monkeypox related claims. Leveraging information from existing fact checking sources and published World Health Organization (WHO) information, we created an open-sourced corpus of 225 rated monkeypox claims. Additionally, we trained an open-sourced BERT-based machine learning model for specifically classifying monkeypox information, which achieved 96\% cross-validation accuracy. PoxVerifi is a Google Chrome browser extension designed to empower users to navigate through monkeypox-related misinformation. Specifically, PoxVerifi provides users with a comprehensive toolkit to assess the veracity of headlines on any webpage across the Internet without having to visit an external site. Users can view an automated accuracy review from our trained machine learning model, a user-generated accuracy review based on community-member votes, and have the ability to see similar, vetted, claims. Besides PoxVerifi’s comprehensive approach to claim-testing, our platform provides an efficient and accessible method to crowdsource accuracy ratings on monkeypox related-claims, which can be aggregated to create new labeled misinformation datasets.

\end{abstract}


\section{Introduction}
Monkeypox is a zoonotic virus related to the virus that causes smallpox. It was first described in humans in 1970 \cite{ladnyj_human_1972, thornhill_monkeypox_2022}. Though it has been traditionally restricted to particular regional outbreaks, on July 23, 2022, the World Health Organization (WHO) declared monkeypox a Public Health Emergency of International Concern \cite{world_health_organization_who_2022}. As of August 5, 2022, the US Centers for Disease Control and Prevention (CDC) reported monkeypox had spread to 88 total countries with 28,220 confirmed cases \cite{centers_for_disease_control_2022_2022}. 

Shortly after monkeypox cases appeared across the globe \cite{kluge_monkeypox_2022}, misinformation related to monkeypox began circulating widely on the Internet and contributing to inaccurate health information \cite{ryan_monkeypox_2022,ennab_monkeypox_2022}. Misinformation surrounding previous monkeypox outbreaks has led to negative health outcomes (e.g., stigma which increased delays in seeking treatment and suppressed reporting to public health authorities which undermined education efforts and increased infection risk \cite{brainard_misinformation_2020}). The ability of misinformation to cause stigma \cite{ennab_monkeypox_2022,brainard_misinformation_2020,hatzenbuehler_prospective_2011, herek_hiv-related_2002}, as occurred with COVID-19 misinformation\cite{islam_covid-19related_2020}, is also of particular concern during monkeypox \cite{williams_homophobic_2022}, which further undermines effective response strategies \cite{unaids_unaids_2022}. Widely disseminated disinformation that claims monkeypox only affects gay and bisexual men could further marginalize LGBTQ+ communities \cite{hart_monkeypox_2022, migdon_experts_2022}. Given the unprecedented speed and spread of health misinformation on social media more broadly \cite{vosoughi_spread_2018} and considering the harms caused by xenophobic, inaccurate characterizations during the HIV/AIDS epidemic 1980s \cite{hatzenbuehler_stigma_2013,world_health_organization_immunizing_2020, lewandowsky_misinformation_2012, lewandowsky_countering_2021}, experts quickly recognized and warned the public about the dangers of falsely believing that monkeypox only threatens sexually-active gay men \cite{boghuma_monkeypox_2022}. 

Due to the continued dearth of available monkeypox misinformation tools currently available, we developed PoxVerifi, a Google Chrome extension used to help users verify the accuracy of monkeypox-related articles by 1) providing automated accuracy reviews 2) displaying similar vetted claims to compare with and 3) enabling users to vote on the accuracy of claims and see what others in the community have voted. PoxVerifi is publicly available to download at no cost on the Chrome Store\footnote{\href{https://chrome.google.com/webstore/detail/poxverifi/ddjpfhchdnofmaecejpmemaldpofmgoi}{chrome.google.com/webstore/detail/poxverifi/ddjpfhchdnofmaecejpmemaldpofmgoi}}. Our code is available as open-source on GitHub\footnote{\href{https://github.com/computationalmedialab/PoxVerifi}{github.com/computationalmedialab/PoxVerifi}} to enable future work to utilize and expand the tool. A video demonstration of  PoxVerifi, including installation instructions and usage is available on YouTube\footnote{\href{https://youtu.be/cCu-4Divdbg}{youtu.be/cCu-4Divdbg}}.

\subsection{Other Monkeypox-related Misinformation Studies and Tools}

The emerging literature on monkeypox-related misinformation reveals worrying trends that are reminiscent of both the COVID-19 pandemic and HIV epidemic. For example, nearly half of the US public report that they have limited knowledge about monkeypox \cite{malik_attitudes_2022}, which exposes a collective uncertainty about the disease. Uncertainty is generally an aversive state which motivates people to seek information that reduces unknowns \cite{webster_individual_1994}. COVID-19-related uncertainty rendered people more vulnerable to misinformation \cite{larsen_conspiratorial_2021}. Indeed, the COVID-19 pandemic illustrated the harms of spreading disinformation at a time of high collective public uncertainty. Many public health officials are alarmed by xenophobic undertones which are driving current misinformation surrounding monkeypox and undermining public health remedies  \cite{boghuma_monkeypox_2022}.

Research has started to map the scale and impact of monkeypox misinformation across various platforms. Most notably, Ortiz-Martínez et al. found that 52\% of monkeypox-related tweets contained misinformation or unverifiable information and that these misleading tweets were more likely to receive more replies, retweets, and likes than truthful tweets \cite{ortiz-martinez_monkeypox_2022}. Other work found that 11.9\% of highly viewed YouTube monkeypox-related videos were also misinformative \cite{ortiz-martinez_youtube_2022}. 

Informed by the potential of misinformation to adversely affect public health, researchers have begun collecting publicly available monkeypox-related social media content in an effort to preemptively counter the misinformation. For example, Thakur \cite{thakur_monkeypox2022tweets_2022} and Movahedi Nia et al. \cite{movahedi_nia_twitter_2022} built datasets of over 255,000
and 800,000 tweets related to 2022 monkeypox outbreaks respectively. There are also emerging development projects on the software repository GitHub with tools to specifically target monkeypox misinformation such as a transformer model\footnote{\href{https://huggingface.co/smcrone/monkeypox-misinformation
}{huggingface.co/smcrone/monkeypox-misinformation
}} trained on a labeled Twitter dataset\footnote{\href{https://kaggle.com/datasets/stephencrone/monkeypox}{kaggle.com/datasets/stephencrone/monkeypox}} to classify monkeypox misinformation \cite{crone_monkeypox_2022}. This work represents a vanguard. As such, there remains a dearth and further work in these areas is needed.

\subsection{PoxVerifi Features and Data Availability}
We combined strategies of past misinformation detection work \cite{botnevik_brenda_2020, kolluri_coverifi_2021, pandey_machine_2020} to rapidly build a scalable tool to identify and curtail monkeypox misinformation in circulation to mitigate harms to public health. Some of the key aspects and features of PoxVerifi are detailed in sections 1.2.1-1.2.5.
\subsubsection{PoxVerifi Misinformation Dataset}
We created a dataset of labeled monkeypox information consisting of 170 factual claims and 55 false claims collected from existing fact-checking repositories and World Health Organization articles on monkeypox. These data are available as open-source on Github.\footnote{\href{https://github.com/computationalmedialab/PoxVerifi/tree/master/Datasets}{github.com/computationalmedialab/PoxVerifi/tree/master/Datasets}}

\subsubsection{Crowdsourced Voting}
We extend Kolluri’s \& Murthy’s \cite{kolluri_coverifi_2021} system of crowdsourced voting to PoxVerifi. Specifically, PoxVerifi allows users to vote on whether or not they think a claim is credible and can later see what other users have voted. We further improve upon this feature by 1) allowing the ability to vote on claims across multiple platforms on the Internet (e.g., news sources, blogs, etc.) without leaving the page and 2) showing users similar, vetted claims to make the most informed decisions possible. 

\subsubsection{Holistic User Experience}
PoxVerifi does not require users to visit an external website in order to verify claims. PoxVerifi combines user generated accuracy reviews, machine-generated accuracy reviews, and the ability to view similar claims. This is all done on a single pop-up that leads to no disturbances of a news viewing experience.

\subsubsection{Labeled, Scalable Monkeypox Claims Data}
Because PoxVerifi allows users to provide an accuracy rating on articles, PoxVerifi can be successfully scaled to crowdsource larger datasets of labeled data, thus providing an open-sourced database (with claims labeled as true or false), available to researchers facilitating textual or content analyses, etc. on misinformation around monkeypox.

\subsubsection{Bias Reduction}
While individuals (including experts) routinely make errors in judgments \cite[p.83]{kahneman_thinking_2013}, PoxVerifi employed two approaches to reduce bias and increase the accuracy of the pooled judgments of voters. First, users can access similarly vetted claims, which can help users overcome heuristic biases by activating deliberate cognitive processing of information. Deliberate cognitive processing is better for making complex judgments because it is more analytical and effortful \cite[p.187]{ormrod_human_2018}. Second, users can only see how other individuals voted after they cast their ballot to avoid the bias of peer pressure and voting as others did.


\section{Software Tools}
\subsection{Frameworks and Hosting Tools}
\subsubsection{Front-end}
The front-end of PoxVerifi was developed in React.js\footnote{\href{https://reactjs.org/}{reactjs.org}}, a JavaScript library used traditionally to create websites. The front end does not require a hosting platform because Chrome extensions are run locally on users' devices. 

\subsubsection{Back-end}
The back-end of PoxVerifi is composed of three parts: a proprietary machine learning model to detect misinformation, a fuzzy search algorithm for claim matching, and a database to store user generated votes. The machine learning model and fuzzy search algorithm were created with Flask\footnote{\href{https://flask.palletsprojects.com/en/2.2.x/}{flask.palletsprojects.com/en/2.2.x/}}—a Python-based framework to develop API endpoints. The Flask API is hosted on a Oracle Cloud Compute Instance. The database to store user votes is a Cloud Firestore hosted on Google Firebase and allows for up to 1GiB of storage, 50,000 reads per day, and 20,000 writes per day on a free plan.

\subsection{ Vetted Monkeypox-Related Information }
When a user visits a website, PoxVerifi shows the user a set of vetted claims which are textually similar to the headline of the website being viewed. To do this, we created a database of accuracy-rated claims related specifically to monkeypox. We used the Google Fact Check API\footnote{\href{ https://toolbox.google.com/factcheck/apis}{toolbox.google.com/factcheck/apis}}, which aggregates claims from many reliable fact-checking sources, including Politifact and FactCheck.org. Using this API, we queried the term, “monkeypox,” and filtered out any non-English claims. We limited our search to “monkeypox” only to prevent potential noise that could come from more general terms such as “pox.” Additionally, the Google Fact Check API employs a topic tagging function, which automatically tags claims related to monkeypox, and, thus, our query returned all of the tagged claims. Claims which had sentence-length ratings which were collapsed to a single phrase representing the ratings so they could be succinctly represented on the website (e.g., “There is no evidence this is true. Monkeypox is caused by a virus.” was changed to “no evidence”). At the time of creation, the resulting corpus contained 56 accuracy-rated claims that we were able to extract from this process, likely due to a recent flurry of monkeypox outbreaks. This dataset, however, had only a single claim labeled as true.

To expand our corpus to more true claims we scraped claims directly from articles published by the WHO about monkeypox. We queried every article published by the WHO using the term “monkeypox,”\footnote{\href{https://who.int/home/search?indexCatalogue=genericsearchindex1\&searchQuery=monkeypox}{who.int/home/search?indexCatalogue=genericsearchindex1\&searchQuery=monkeypox}} and then scraped the contents of each of the 90 articles returned. Each article was then parsed into individual sentences by splitting the article body by sentence-ending punctuation (e.g., “.”,”?”,”!”). Because not every sentence contains a claim with a verifiable fact, we used the ClaimBuster API, which assigns a score between 0 and 1 indicating the likelihood that the sentence contains a verifiable claim \cite{hassan_toward_2017}. All articles’ contents were fed into ClaimBuster’s Claim Detection API. We only retained sentences with a score over 0.8; Hassan et al. reported a threshold of 0.5 for “highly check-worthy factual claims,” \cite{hassan_toward_2017} but we raised our threshold to 0.8 to avoid potential noise. Though any monkeypox-related text can contain errors, claims published by the WHO are highly authoritative and the organization has an established track record for authoritative health information on many diseases (e.g, a previous monkeypox outbreak \cite{world_health_organization_world_2011} and dengue fever \cite{world_health_organization_dengue_2009}). As a result, we added the resulting 169 claims to our corpus and labeled them as true. This dataset is freely available as open-source for others\footnote{\href{https://github.com/computationalmedialab/PoxVerifi/tree/master/Datasets}{github.com/computationalmedialab/PoxVerifi/tree/master/Datasets}}. 

\subsection{Machine Learning Model}
Various approaches to machine learning has been applied by others to identify aspects of monkeypox-related information circulating on the Internet. For example, Mohbey et al. \cite{mohbey_cnn-lstm-based_2022} created a proprietary hybrid deep learning approach in order to classify the sentiment of tweets as positive, negative, or neutral. Sv and Ittamall \cite{sv_what_2022} used a pre-trained sentiment analysis library to find sentiment distribution of monkeypox-related tweets and used Latent Dirichlet Allocation (LDA), an unsupervised generative probabilistic topic modeling method first introduced by Blei et al. \cite{blei_latent_2003}. LDA is among the most common topic modeling methods \cite{jelodar_latent_2019}.

However, as the scope of these studies suggests, machine learning has not been extensively applied to the specific task of identifying monkeypox-related misinformation. Rather, machine learning has been more focused on the diagnosis of the disease \cite{tom_neuro-fussy_2018}. To address this gap in the literature, we created our own approach to detect monkeypox-related misinformation. Specifically, we trained a Bidirectional Encoder Representations from Transformers (BERT)-based model on the data we collected to classify monkeypox-related information. First, our vetted data were collapsed into two categories: true and false. We converted the nominal labels of “no evidence”, “false”, “inaccurate”, “mostly false”, “misleading”, “incorrect”, “half true”, “not required”, “unsupported”, “needs context” were converted to just false, and the remaining nominal label of “true” remained as true. Our dataset was made up of claims where 170 were true and 55 were false. We trained a model using the package DistilBERT, a smaller, faster version of BERT \cite{sanh_distilbert_2020}, and extended and developed Alammar’s \cite{alammar_visual_2019} Python notebook.

Since the size of our dataset was limited, we used 10-fold cross-validation—a method that trains and tests 10 independent models on different splits of train and test data and averages the accuracy together—to test our models' accuracy to ensure that our accuracy was representative of our data. Our trained BERT-based model achieved a high level of accuracy on the data set we created: 96\% 10-fold cross-validation accuracy. The codebook used to train the model as well as the saved model parameters are available as open-source on GitHub\footnote{\href{https://github.com/computationalmedialab/PoxVerifi/tree/master/BERT-Model-Training}{github.com/computationalmedialab/PoxVerifi/tree/master/BERT-Model-Training}}.

\section{PoxVerifi Extension Design}
In order to rapidly create an accessible tool that could be used at scale to combat monkeypox-related information, we opted to create a compact, computationally light browser extension made widely available on the Google Chrome Store as a no-cost download. PoxVerifi’s software is made up of three components: 1) a front-end Google Chrome extension 2) a back-end database for live vote storage and 3) a back-end Python framework to parse news sources, compare headlines to previously vetted claims, and provide machine-learned ratings.
\begin{figure}[ht]
\centering
\includegraphics[width=6.5in]{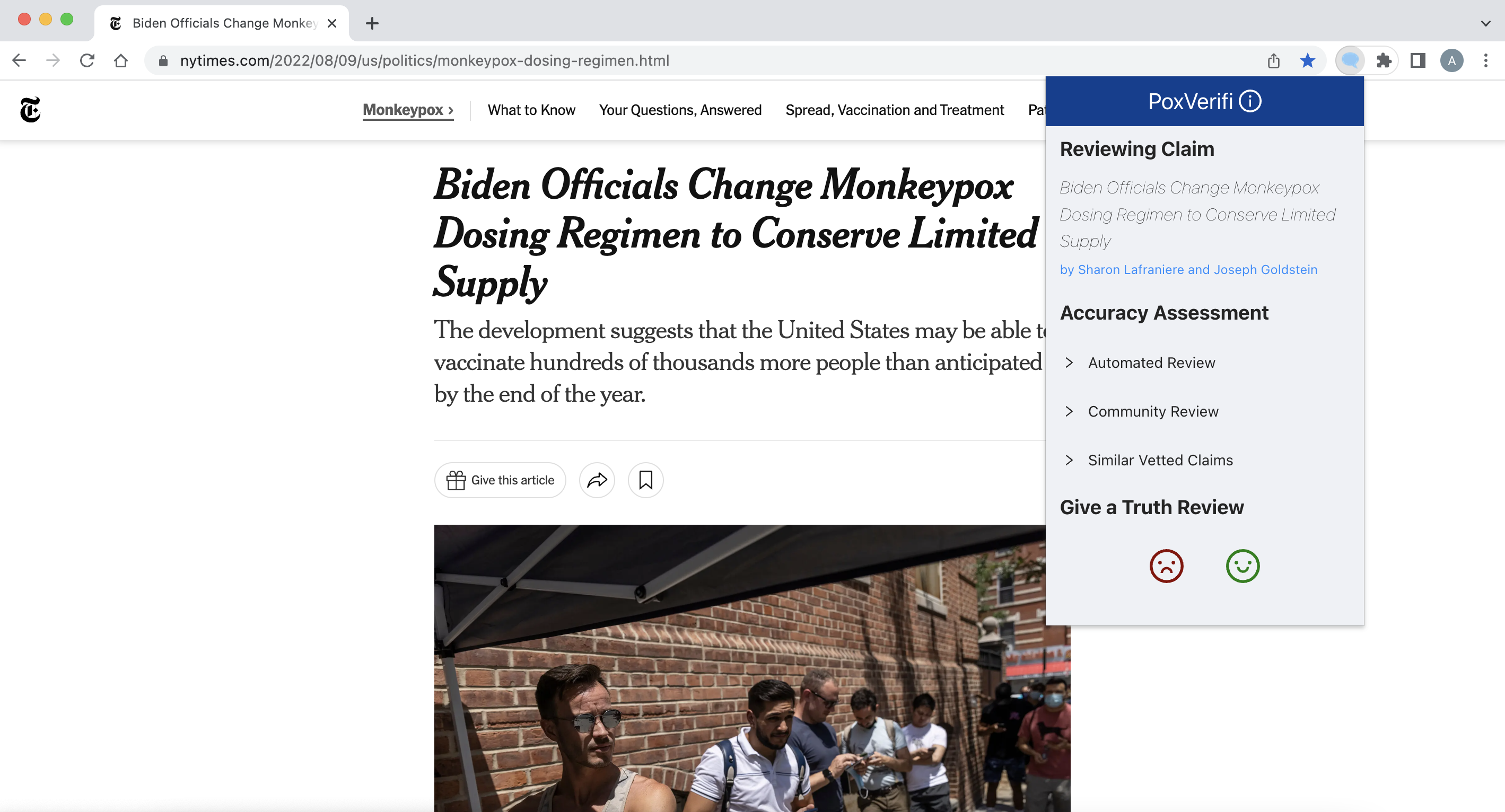}
\vspace{2pt}
\caption{PoxVerifi being used on a sample news article.}
\vspace{6pt}
\end{figure} 

\begin{figure}[!tbp]

\begin{minipage}[b]{0.45\textwidth}
    \includegraphics[width=3.0in]{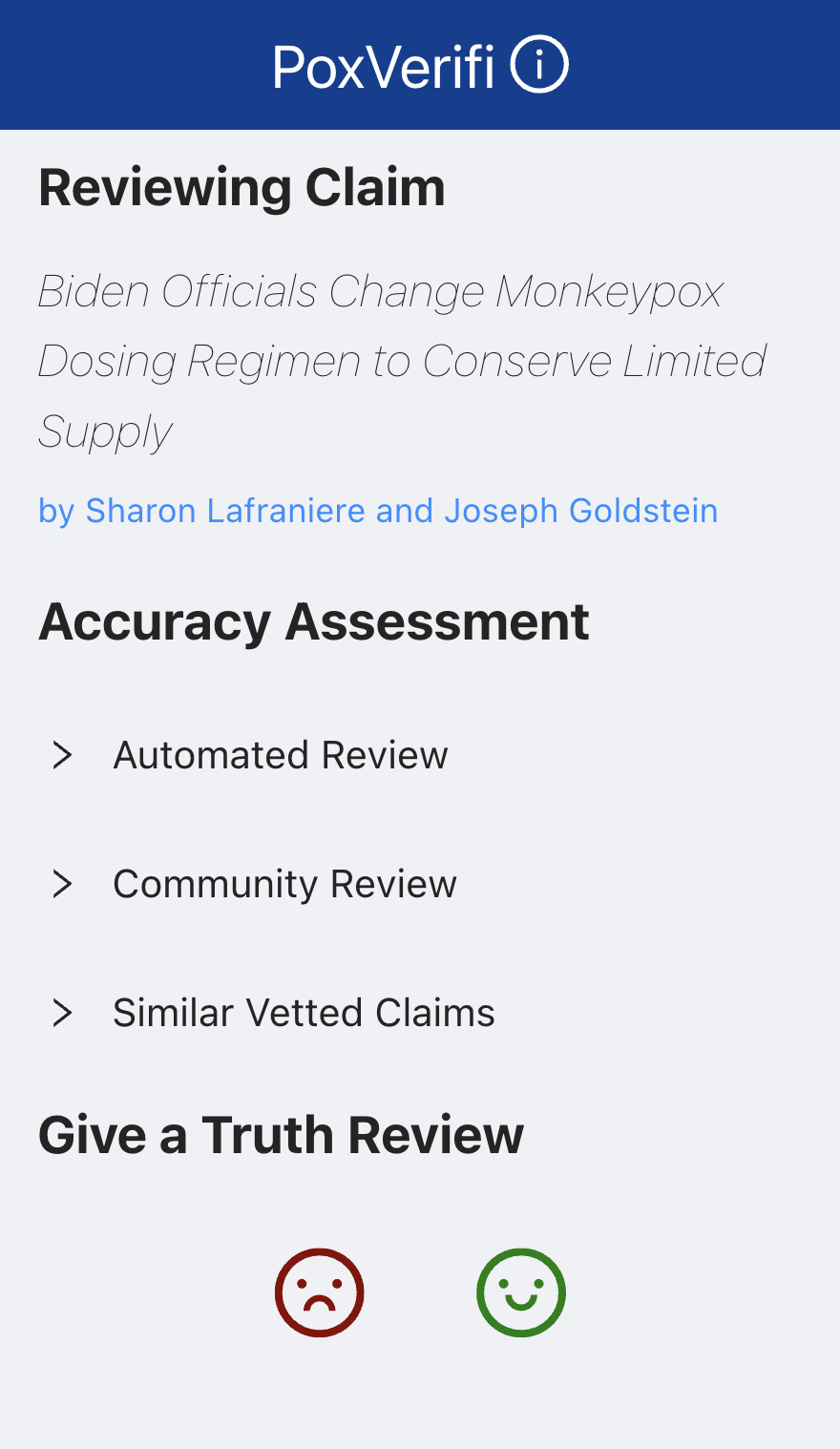}
\caption{PoxVerifi display of a claim for review.}
\vspace{78pt}
     \includegraphics[width=3.0in]{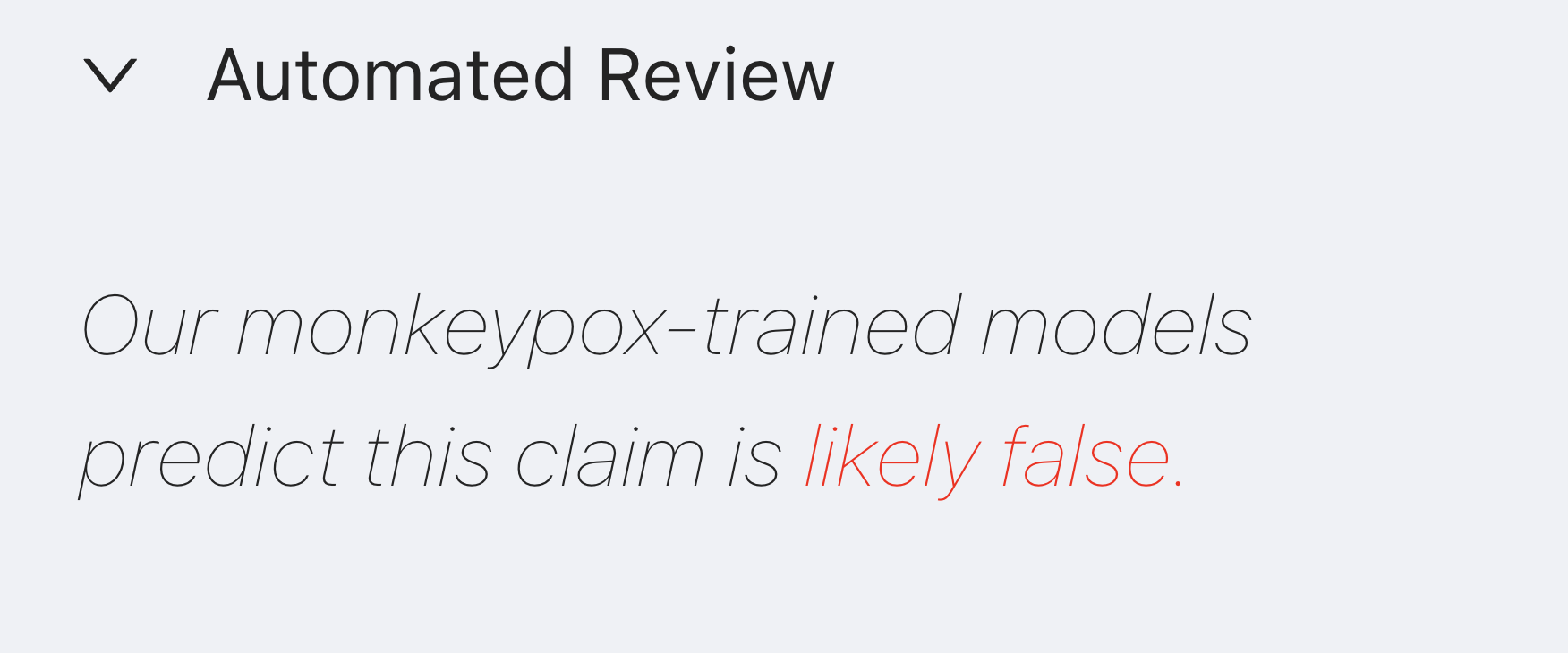}
\caption{Sample accuracy display after expanding the “automated review” section for a claim determined to be false.}

  \end{minipage}
\hfill
\begin{minipage}[b]{0.45\textwidth}
\centering

\includegraphics[width=3.0in]{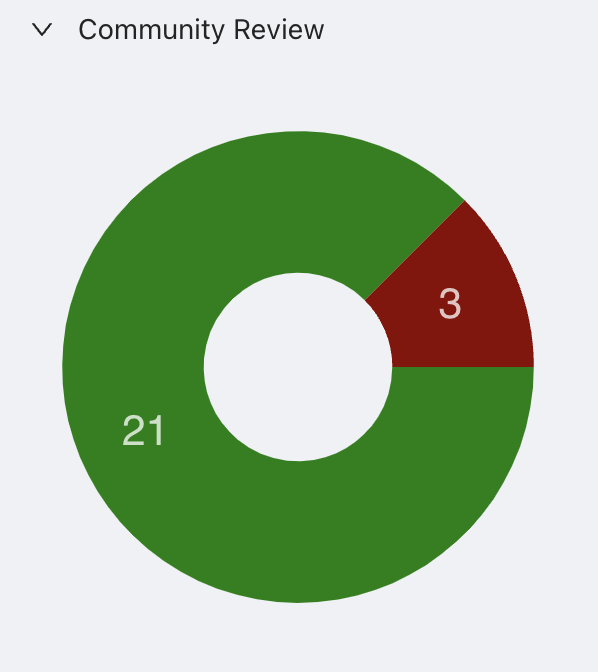}
\caption{Sample pop up after selecting community based-review (only available after casting a vote).}
\vspace{40pt}
\includegraphics[width=3.0in]{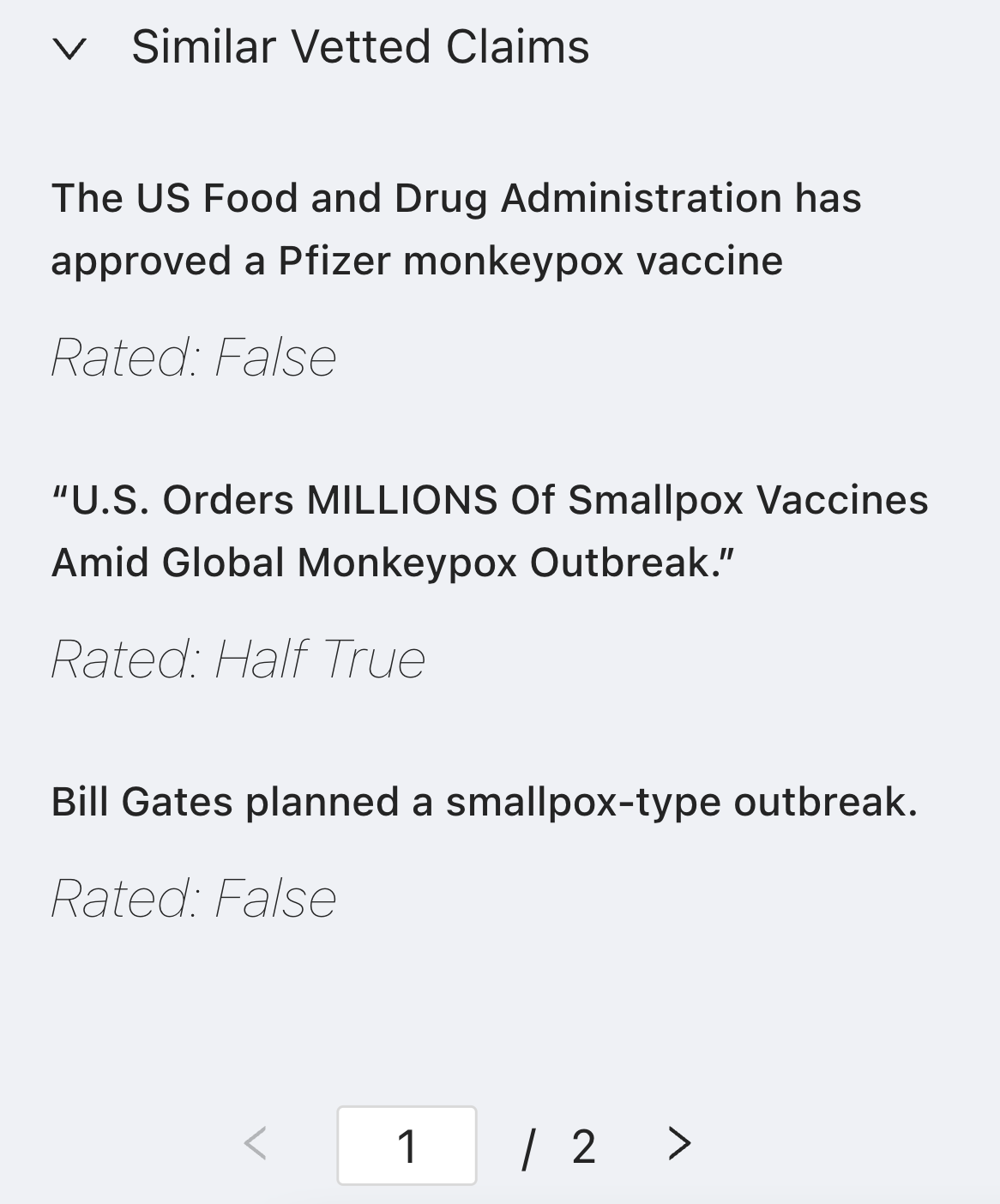}
\caption{Sample claim display after user expands the “similar vetted claims” section. }

  \end{minipage}
\end{figure} 

\subsection{Chrome Extension Interface}

At any point during a browsing session, users can click on the PoxVerifi extension icon in the top right corner of their browser. Once a user encounters a claim on the Internet that they wish to verify, they click the extension and a loading screen is displayed. While the loading screen is visible, the URL of the current website, containing the monkeypox claim, is sent to the backend headline\_detection API method. If there is no headline detected (e.g., there is only an image or it is the homepage of a browser), the extension will instruct the user to view a news source. If a headline is detected, it will be displayed along with the author of the article. PoxVerifi then analyzes the accuracy of the displayed headline.

Below the headline section (see figure 2), is a section titled “Accuracy Assessment" with three drop downs. When clicked, they display: 1) ML Model Accuracy Estimates (see figure 3) 2) Other Users Crowdsourced votes (see figure 4) and 3) Similar Vetted Claims (see figure 5). Finally, at the bottom is a section enabling users to cast a vote. 

\subsubsection{ML Accuracy Estimate}
The news headline is sent to the backend ml\_classification API method which returns a parameter representing the accuracy determined by our machine learning model, which is displayed for the user to see.

\subsubsection{Crowdsourced User Votes}
A call is made to the Firebase database using the news URL as a key, and past users’ votes are retrieved. This is displayed in a circular pie graph, displaying red, yellow, and green sections corresponding to fake, mixed, and truthful reviews (which labels appear on hover). Crowdsourced votes can only be viewed after the user casts their own vote, to avoid the pull of conformity influencing a user’s vote \cite{guo_conformity_2019}.

\subsubsection{Similar Vetted Claims}
The news headline is sent to the backend get\_similiar\_claims API method, which returns a list of all vetted claims similar to the headline and their corresponding review. Pagination allows users to click through to continue viewing claims until they have enough context/have found a similar claim to make an informed decision.

\subsubsection{Vote Casting}
Users can cast a vote to review the truth of the claim by clicking one of two buttons: inaccurate or misleading news (red, frowning emoji) or accurate news (green, happy face). After voting, a confirmation message is displayed affirming their vote, with the opportunity to revoke the vote if the user made a mistake. In order to prevent abuse of the vote casting system, each installation of the Google Chrome extension is limited to only voting one time at each unique URL (see the bottom of Figure 2 under “Give a Truth Review” for a visual depiction).

\subsection{Backend}

Our Flask-based backend has three API endpoints, all accessible using an HTTP GET request. 

\subsubsection{headline\_detection}
This function takes a parameter of a website URL. This website URL is used to query the content of the webpage article. This is then parsed, and the headline of the webpage is returned.

\subsubsection{ml\_classification}
This function takes a parameter of a headline. The news headline first goes through a preprocessing step to make sure that it has a claim for our model to check. New headlines are fed into a machine learning model as described in section 2.3, which produces a probability that can be coded as a binary prediction. A nominal value of 0 or 1 is returned, where 0 represents a fake determination and 1 represents a true determination.

\subsubsection{get\_similar\_claims}
This function takes a parameter of a headline. The Python package fuzzysearch\footnote{\href{https://pypi.org/project/fuzzysearch/}{pypi.org/project/fuzzysearch/}} is used to generate a similar score (based on the number of approximately the same words) between the parameter headline, and the headlines the other etted scraped claims scraped as described in section 2.2. All claims which have a similarity score with the parameter claim of over 50 (which we observed was a sufficient threshold to yield topically similar sentences) are returned.

\subsubsection{Database}
PoxVerifi’s database, hosted on Cloud Firestore, is made up of documents each identified by the URL of a news article. Each document record contains a fake, neutral/mixed, and true vote count.

\section{Discussion}
PoxVerifi is built with an extensible framework to accurately and conveniently verify the veracity of monkeypox-related claims. By integrating features of other misinformation detection tools \cite{botnevik_brenda_2020,kolluri_coverifi_2021,pandey_machine_2020}, PoxVerifi provides a reliable methodology for testing claims. We successfully trained a deep machine learning model on monkeypox-related information that achieves 96\% cross-validation accuracy on classifying news as likely true or likely false. 

Our tool also allows users to access previously vetted claims to compare the claim. This provides a reference point and the act of deliberating and comparing claims engages active, deliberative processing. Only after voting on the truth of a claim can users see how others voted. This reduces systemic bias in the data by ensuring that each vote is made independent of the existing group results. This is important in reducing the human tendency to follow the crowd, which can render one’s judgment more vulnerable to errors based on overweighting of partial or false evidence \cite[p.89]{cialdini_influence_2006}. This will help ensure that each vote is cast independently of others and is crucial to ensure the accuracy of crowdsourced voting. The judgment errors by individuals that contribute to group data tend to average to zero as long as there is no hidden systemic biases (e.g., seeing how the majority voted before choosing and that observations are independent and their errors are uncorrelated \cite[p.84]{kahneman_thinking_2013}). PoxVerifi meets these conditions.

As a Google Chrome extension, PoxVerifi can be used by almost anyone\footnote{Anyone who has a computer and software that can run Chrome and access the Chrome Extension Store.
} with Internet access to rate claims and view the ratings of claims. By using a variety of sources, PoxVerifi can help mitigate monkeypox-related misinformation by providing a highly accurate, easy-to-use, accessible tool to assess the veracity of almost any claim. 

PoxVerifi also has the potential to rapidly and inexpensively yield crowdsourced labeled datasets if it is well adopted. While user votes are not always aligned with automatic models, cases where they do align could indicate high confidence in labeling. As more people install PoxVerifi and rate articles, these crowdsourced votes can be used to automatically create highly accurate labels for new sources. This labeled data can be leveraged in future work to study monkeypox-related misinformation. 

PoxVerifi addresses a key concern about fact-checking aids. Namely, by giving users access to previously-vetted claims, we reduced the potential for users to rely solely on heuristic cognitive strategies tied to an implicit trust of automatically-generated credibility labels. In a study of detecting misinformative images \cite[p.39]{giotta_ways_2020}, providing credibility labels was found to reduce the amount of human deliberation. PoxVerifi is designed to engage deeper processing of information in general, activating cognitive processes that enable users to become more cognizant of misinformation. By showing users similar claims that have been previously vetted and labeled, users can more easily access the cognitive processes required to make evidence-based comparisons when determining the accuracy of claims they encounter in news stories. Previous work found that  truth nudging resulted in a 3-fold increase in users’ ability to discern truthful headlines relative to false headlines \cite{pennycook_fighting_2020}. PoxVerifi provides a framework to facilitate similar studies in the context of monkeypox claims.

\section{Limitations and Future Work }
\subsection{Limitations}
A limitation of our work is partially due to the lack of accessible monkeypox-related claims at the time we developed the PoxVerifi tool. Indeed, the dearth of high-quality, labeled monkeypox-related information poses some inescapable limitations. After exhaustively aggregating multiple sources, we were only able to achieve a total of 225 pieces of content. Although we were still able to develop a model achieving high accuracy (e.g., 96\%), the representativeness of the training data is somewhat limited. Moreover, to err on the side of accuracy, we only used WHO content to ensure the highest level of veracity of claims. However, future models could be improved by incorporating a variety of highly credible sources to increase data representativeness. Yet, given the rapidly changing nature of public health emergencies, this is a common obstacle. However, as PoxVerifi accrues more data through crowdsourcing, it will generate new data points that can be used in the future to train more accurate machine learning models.

\subsection{Future Work}
Future work includes scaling PoxVerifi to be able to automatically collect new information from existing fact-checking repositories in order to keep all information up-to-date. Furthermore, while PoxVerifi’s machine learning model does not currently continuously train as users label more data, future work can be done to implement a continuous learning feature for our ML model based on user-generated votes. Future work can also extend our PoxVerifi framework to other use cases such as the seasonal flu, childhood diseases, and the COVID-19 pandemic. These could be quickly adapted and scaled to mitigate misinformation harms. 

\section{Conclusion}
The circulation of monkeypox-related misinformation is perpetuating harmful ideas that negatively impact response strategies. Misinformation about monkeypox can marginalize LGBTQ+ communities and could exacerbate the severity of the outbreak \cite{unaids_unaids_2022,migdon_experts_2022, hatzenbuehler_stigma_2013, boghuma_monkeypox_2022}. It is therefore imperative to quickly develop tools that can combat monkeypox-related misinformation and disinformation. To help address this need, we introduced PoxVerifi, an open-source Google Chrome extension that enables people to develop more accurate monkeypox-related news consumption habits. PoxVerifi uses a comprehensive set of checks, including automated reviews, community reviews, and the ability to view similar claims to help people better understand the credibility of news they are reading. As a Google Chrome extension, PoxVerifi does all of this without requiring a user to visit external websites, which facilitates relatively uninterrupted news reading experiences. PoxVerifi also provides an open source framework to quickly obtain labeled misinformation data through crowdsourcing. This enables future work to build upon the PoxVerifi codebase to create more advanced crowdsourced information tools to tackle other types of misinformation, health-related or otherwise.

\section*{Acknowledgments}

This work was supported by Good Systems, a research Grand Challenge at the University of Texas at Austin. This work was supported in part by Oracle Cloud credits and related resources provided by the Oracle for Research program. We thank Nikhil Kolluri for providing suggestions of features to include in PoxVerifi and River Terrell for providing manuscript feedback.

 \printbibliography

\end{document}